\documentclass{article}

\PassOptionsToPackage{numbers}{natbib}

\usepackage[preprint]{neurips_2023}

\usepackage[T1]{fontenc}    
\usepackage{url}            
\usepackage{booktabs}       
\usepackage{amsfonts}       
\usepackage{nicefrac}       
\usepackage{microtype}      
\usepackage[table,usenames,dvipsnames]{xcolor}      
\usepackage{xcolor}
\usepackage{colortbl} 

\definecolor{aliceblue}{gray}{0.1}
\usepackage[colorlinks,linktoc=all]{hyperref}
\hypersetup{citecolor=Blue}
\hypersetup{linkcolor=Blue}
\hypersetup{urlcolor=Blue}

\usepackage{hyperref}

\usepackage{amsmath}
\usepackage{amssymb}
\usepackage{mathtools}
\usepackage{amsthm}
\usepackage{xspace}

\usepackage{amsmath,amsfonts,bm}

\def\eqref#1{equation~\ref{#1}}

\def\1{\bm{1}}

\DeclareMathAlphabet{\mathsfit}{\encodingdefault}{\sfdefault}{m}{sl}
\SetMathAlphabet{\mathsfit}{bold}{\encodingdefault}{\sfdefault}{bx}{n}

\usepackage[capitalize,noabbrev]{cleveref}

\theoremstyle{plain}

\theoremstyle{definition}

\theoremstyle{remark}

\usepackage[textsize=tiny]{todonotes}

\usepackage[textsize=tiny]{todonotes}
\usepackage{multirow}
\usepackage{enumitem}

\title{AnyTaskTune: Advanced Domain-Specific Solutions through Task-Fine-Tuning} 

\author{DataTager\thanks{Please cite this work as ``DataTager(2024)". Full authorship contribution statements appear at the end of the document. Correspondence regarding this technical report can be sent to \url{report@datatager.com}}}

\usepackage[subtle]{savetrees}

\begin{document}
\maketitle

\begin{abstract}
The pervasive deployment of Large Language Models-LLMs in various sectors often neglects the nuanced requirements of individuals and small organizations, who benefit more from models precisely tailored to their specific business contexts rather than those with broadly superior general capabilities. This work introduces \textbf{AnyTaskTune}, a novel fine-tuning methodology coined as \textbf{Task-Fine-Tune}, specifically developed to elevate model performance on a diverse array of domain-specific tasks. This method involves a meticulous process to identify and define targeted sub-tasks within a domain, followed by the creation of specialized enhancement datasets for fine-tuning, thereby optimizing task-specific model performance. We conducted comprehensive fine-tuning experiments not only in the legal domain for tasks such as keyword extraction and sentence prediction but across over twenty different sub-tasks derived from the domains of finance, healthcare, law, psychology, consumer services, and human resources. To substantiate our approach and facilitate community engagement, we will open-source these bilingual task datasets. Our findings demonstrate that models fine-tuned using the \textbf{Task-Fine-Tune} methodology not only achieve superior performance on these specific tasks but also significantly outperform models with higher general capabilities in their respective domains.
Our work is publicly available at \url{https://github.com/PandaVT/DataTager}.

\vspace{-0.4em}
\end{abstract}

\section{Introduction}

Recently, the rapid advancement and deployment of Large Language Models (LLMs) have transformed various sectors by providing unprecedented natural language processing capabilities~\cite{zhao2023survey}. Traditionally, these models have been developed with a focus on enhancing their general abilities, aiming to create universally powerful tools that excel across a broad spectrum of tasks and domains. However, this generalized approach often fails to meet the specific requirements of individual users and small organizations, whose needs are intricately tied to their unique operational contexts. To address this issue, one very simple way is to fine-tune LLMs for different tasks.
In the medical field, researchers have optimized LLMs using medical datasets to support functionalities such as medical document interpretation and diagnostic consultations. For example, ChatMed~\cite{zhu2023ChatMed}, DISC-MedLLM~\cite{bao2023discmedllmbridginggenerallarge}, HyKGE~\cite{jiang2024hykgehypothesisknowledgegraph}, IvyGPT~\cite{wang2023ivygpt}, and HuatuoGPT~\cite{zhang2023huatuogpt} are notable examples, demonstrating significant advancements within their specialized medical domains compared to generic LLMs. Similarly, in the legal field, researchers have introduced fine-tuned LLMs trained on legal corpora to support activities such as legal research, contract analysis, and legal document summarization. Models like ChatLaw~\cite{cui2024chatlawmultiagentcollaborativelegal}, LawGPT~\cite{zhou2024lawgpt}, and DISC-LawLLM~\cite{yue2023disc} exemplify this trend, demonstrating a deeper grasp of legal language and principles compared to their generic counterparts. Within finance, researchers are leveraging diverse financial datasets, including conversational data and market reports, to fine-tune LLMs for applications such as financial forecasting, risk assessment, and fraud detection. Examples include FinGPT~\cite{yang2023fingpt}, DISC-FinLLM~\cite{chen2023disc}, and PIXIU~\cite{xie2023pixiu}. Additionally, in other fields, models like FaiMA~\cite{yang2024faimafeatureawareincontextlearning}, designed for Multi-domain applications, and Kuaiji~\cite{luo2024kuaijichineseaccountinglarge}, tailored for accounting tasks, illustrate the versatility and potential of fine-tuned LLMs in specialized domains.

To address this discrepancy, we introduce a novel fine-tuning paradigm specifically designed for Explicit Data Sets, which we term "task fine tune." This method diverges from conventional training techniques by emphasizing precision and specificity over general performance. By focusing on Explicit Data Sets that contain clear, directive input-output pairs with specific instructions, this approach allows for precise tailoring of models to perform designated tasks effectively. Such fine-tuning not only enhances the model’s ability to execute particular functions but also significantly improves its applicability and efficiency in real-world scenarios.

The cornerstone of this work, "AnyTaskTune," is to validate and elaborate on the "task fine tune" method, utilizing Explicit Data Sets across various domains. This method involves identifying the spectrum of tasks required in different fields and scenarios and meticulously developing numerous Explicit Data Sets to conduct task-specific fine-tuning. The aim is to optimize the model's performance for individual and corporate use, ensuring that it meets the nuanced demands of users in fields such as finance, healthcare, law, psychology, consumer services, and human resources.

To showcase the effectiveness and versatility of this approach, we conducted experiments using over twenty different Explicit Data Sets that we will open-source. These datasets are bilingual and designed to cater to a diverse global audience, providing a valuable resource for the community to engage in further research and enhance application-specific model improvements.

By concentrating on specific, clearly defined tasks rather than general capabilities, "AnyTaskTune" significantly outperforms traditional models that prioritize breadth over depth. This strategic shift towards more specialized, task-oriented model training represents a critical evolution in how we develop and deploy LLMs, making them more relevant and valuable in industry-specific applications.

The primary contributions of this paper can be summarized as follows:
\begin{itemize}
    \item Introduction of the "task fine tune" paradigm, specifically designed for Explicit Data Sets, emphasizing precision and specificity over general performance.
    \item Development and utilization of over twenty bilingual Explicit Data Sets across various domains to validate the effectiveness of the "task fine tune" method.
    \item Evaluation of multiple models and datasets across various domains, demonstrating that task-specific fine-tuning yields better results.
    \item Open-sourcing of the developed Explicit Data Sets, providing a valuable resource for the community to engage in further research and enhance application-specific model improvements.
    \item Implementation of our data processing model, DataTager-LLM, as an online accessible website, \url{https://datatager.com}.
\end{itemize}

\begin{figure}[htp]
 \centering
 \includegraphics[width=1.0\columnwidth]{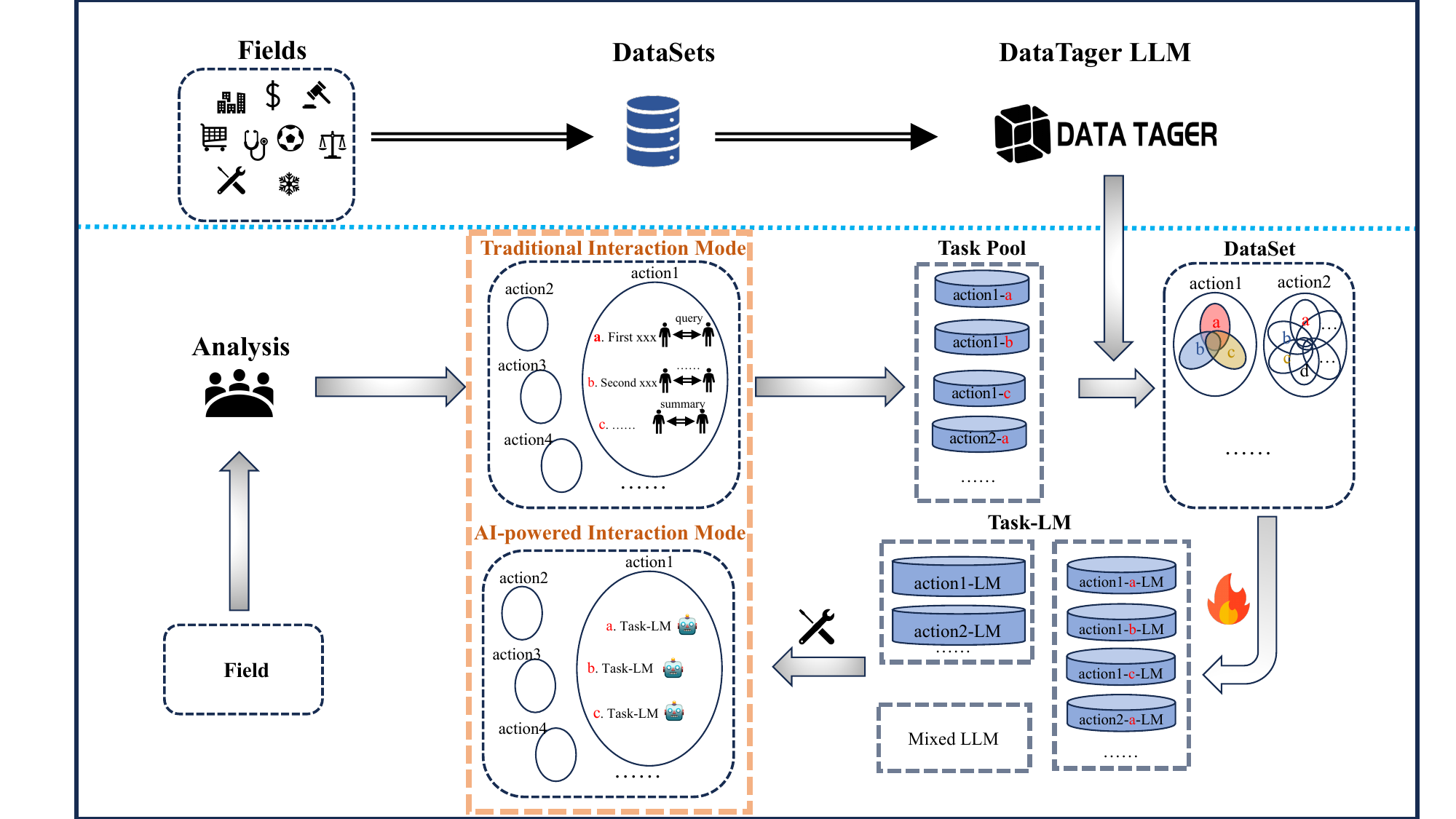}
 \caption{AnyTaskTune Framework}
 \label{fig:motivation2}
\end{figure}

\section{AnyTaskTune}

\subsection{Model Requirements in Business Applications}\label{sec:subtask}
In practical applications, simply pursuing the performance of a general model or deploying a domain-specific model alone is often insufficient to address real-world problems. While scientific research outputs need to be diverse and comprehensive, businesses require standardized and controllable solutions. For most enterprises and organizations, their needs are highly specific and contextualized, which cannot be fully met by a single general language model. In such cases, general models may underperform on specific tasks or fail to achieve the expected efficiency and accuracy. Moreover, while some domain-specific models may provide good performance within their fields, they lack flexibility and scalability, making it difficult to adapt to ever-changing business needs. Therefore, there is a need for a new method to fine-tune and optimize models, ensuring they not only possess the broad applicability of general models but also meet the specific needs of particular domains and tasks. AnyTaskTune addresses this issue by using specially designed explicit datasets for task fine-tuning, ensuring the model's precision and efficiency in specific tasks, thereby enhancing its value in real-world business applications.

\subsection{DataTager}\label{sec:dataset2}
First, leveraging our prior experience and expertise in vertical domains, we collected raw data from multiple fields, including finance, healthcare, law, psychology, consumer services, and human resources. By surveying industry professionals to understand their expectations and requirements for models, we synthesized multiple branch datasets. These datasets cover core tasks across various domains and include detailed and variant data from real-world scenarios. Through the training of these branch datasets, we developed DataTager-LLM, a large-scale data synthesis model. DataTager-LLM forms the foundation of our product and serves as a core tool for further research and application. Currently, we have no plans to open-source DataTager-LLM to better protect our intellectual property and commercial interests.

\subsection{Explicit and Implicit DataSets}\label{sec:dataset3}

\begin{figure}[htp]
 \centering
 \includegraphics[width=1.0\columnwidth]{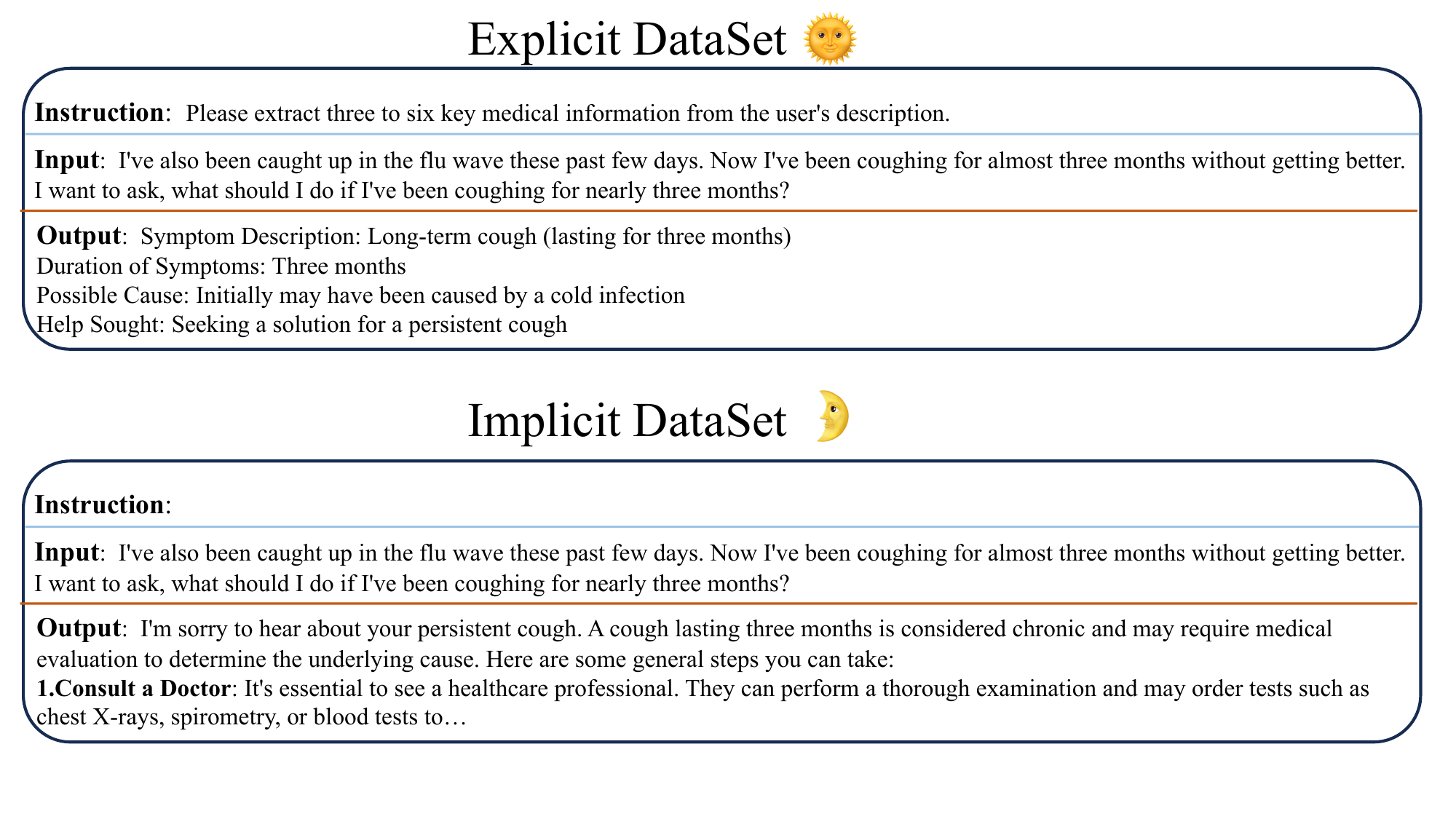}
 \caption{Explicit and Implicit DataSets}
 \label{fig:motivation}
\end{figure}

We defined what constitutes a "good" dataset for businesses and distinguished between these datasets based on the presence of explicit instructions. Specifically, datasets with clear instructions, such as "instruction: Please summarize this news article and extract the key points; input: xxx," are designed for models to perform specific tasks. Explicit datasets provide models with clear guidelines, enhancing their ability to understand and execute specific tasks accurately and efficiently in practical applications. On the other hand, datasets without explicit instructions, such as "instruction: , input: xxx," involve random queries without fixed task types, thus considered implicit datasets. 
Implicit datasets are better suited for handling open-ended questions and unstructured data but are more challenging to train and apply. For businesses looking to enhance their operations with LLMs, the first step is to identify the various task types within their current operational context. This helps determine the number of explicit datasets needed. Each dataset is then synthesized individually, and a general LLM that aligns with the business’s usual interaction patterns is fine-tuned across multiple tasks to create a model that best fits the specific business scenarios.

\subsection{AnyTaskTune}\label{sec:dataset1}

Different fields have multiple interaction (action) modes. Take the medical field, for example: if triage is considered an action, the actual tasks it involves include: a. Patients describing their issues in non-standard language; b. Triage staff translating this into medical standard language; c. Staff analyzing the completeness of the information and asking follow-up questions if necessary; d. Extracting medical key points and directing the patient to the appropriate department. This triage action thus includes four sub-tasks (a, b, c, d). We can use DataTager to generate datasets for each sub-task. These datasets collectively represent the triage action process. Then, we train models based on these datasets, with multiple approaches available:

If we train a separate model for each sub-task, the model can address that specific sub-task effectively. For instance, a model specifically trained for sub-task a can handle the translation of patient descriptions, while another model trained for sub-task b can convert non-standard language into medical standard language.
If we combine these sub-task datasets to train a large mixed model, different instructions will trigger different task types. This method can handle multiple tasks within one model but requires a more complex training process and more computational resources.
It is important to note that if we choose to train multiple small models, based on our experience, a 1.5 billion or 3 billion model can handle less complex tasks well without data drift. If the performance of the mixed model declines, adding some general datasets to the training process can help. This way, we can use high-performing models to replace traditional sub-tasks in action interactions, creating a new interaction model and reducing costs.

\section{Experiments}

To validate the effectiveness and versatility of the AnyTaskTune paradigm, we undertook extensive experiments across multiple domains: finance, healthcare, law, psychology, and role-play. We compared the performance of AnyTaskTune against various models including closed-source large language models (LLMs), open-source LLMs, and domain-specific models. Critically, our experiments maintained a strict separation between training and testing datasets to ensure unbiased evaluation and reproducibility of results.

\subsection{Experimental Setup}

Our experiments were structured as follows:

\begin{itemize}
\item \textbf{Model Base:} We utilized Qwen2-7B~\cite{bai2023qwentechnicalreport} as the base model for AnyTaskTune training. This model was fine-tuned on specific sub-tasks within each domain, and then tested to obtain the experimental results.
\item \textbf{Model Categories:}
\begin{itemize}
\item \textbf{Closed Source LLMs:} GPT-4~\cite{openai2024gpt4technicalreport}, GPT-3.5, LLaMA3-8b, and LLaMA3-70b.
\item \textbf{Open Source LLMs:} Qwen, Baichuan~\cite{yang2023baichuan2openlargescale}, ChatGLM~\cite{glm2024chatglm}, and internalLM~\cite{cai2024internlm2}.
\item \textbf{Domain Models:} ChatMed~\cite{zhu2023ChatMed}, DISC-MedLLM~\cite{bao2023discmedllmbridginggenerallarge}, FinGPT~\cite{yang2023fingpt}, ChatLaw~\cite{cui2024chatlawmultiagentcollaborativelegal}, DISC-LawLLM~\cite{yue2023disclawllm}, MindChat~\cite{MindChat}, SoulChat~\cite{chen2023soulchatimprovingllmsempathy}, and MachineMindset (MBTI)~\cite{cui2024machinemindsetmbtiexploration}.
\end{itemize}
\item \textbf{Evaluated Domains and Tasks:}
\begin{itemize}
\item \textbf{Medical:} Sub-tasks M1, M2, M3.
\item \textbf{Finance:} Sub-tasks F1, F2, F3.
\item \textbf{Law:} Sub-tasks L1, L2, L3, L4.
\item \textbf{Psychology:} Sub-tasks P1, P2.
\item \textbf{Role-Play:} Sub-task MBTI.
\end{itemize}
\end{itemize}

\subsection{Results and Domain-Specific Analysis}
The results of the experiments are summarized in Table~\ref{tab:results}. Our AnyTaskTune models, fine-tuned on specific domain data, demonstrated significant performance improvements, even sometimes surpassing well-regarded models like GPT-4 and LLaMA3-70b. The following points detail cross-domain influences and task-specific adaptabilities observed during testing:

\begin{table}[htbp]
    \centering
    \renewcommand{\arraystretch}{1.15}  
    \setlength{\tabcolsep}{1.1mm}       
    \tiny                       
    \caption{Performance Comparison Across Various Domains and Models}
    \label{tab:results}
    \begin{tabular}{l|ccc|ccc|cccc|cc|c}
        \toprule[1.5pt]
        & \multicolumn{3}{c}{\textbf{Med}} & \multicolumn{3}{c}{\textbf{Finance}} & \multicolumn{4}{c}{\textbf{Law}} & \multicolumn{2}{c}{\textbf{Psychology}} & \textbf{Role-Play} \\
        \cmidrule(lr){2-4} \cmidrule(lr){5-7} \cmidrule(lr){8-11} \cmidrule(lr){12-13} \cmidrule(lr){14-14}
        & \textbf{M1} & \textbf{M2} & \textbf{M3} & \textbf{F1} & \textbf{F2} & \textbf{F3} & \textbf{L1} & \textbf{L2} & \textbf{L3} & \textbf{L4} & \textbf{P1} & \textbf{P2} & \textbf{MBTI} \\
        \noalign{\hrule height 1.5pt}
        \rowcolor{gray!20}\multicolumn{14}{c}{\it{\textbf{Closed Source LLMs}}} \\
        \hline
        GPT4 & 0.526 & \textcolor{blue}{0.671} & \textcolor{blue}{0.643} & \textcolor{blue}{0.492} & 0.615 & \textcolor{blue}{0.719} & 0.592 & \textcolor{blue}{0.593} & 0.636 & 0.662 & \textcolor{blue}{0.521} & \textcolor{blue}{0.661} & \textcolor{blue}{0.731} \\
        GPT3.5 & 0.493 & 0.482 & 0.536 & 0.312 & \textcolor{blue}{0.641} & 0.623 & 0.471 & 0.526 & 0.596 & 0.562 & 0.482 & 0.472 & 0.652 \\
        LLaMA3-8b & 0.267 & 0.389 & 0.419 & 0.316 & 0.523 & 0.462 & \textcolor{red}{0.195} & 0.325 & 0.286 & 0.353 & 0.312 & 0.278 & 0.259 \\
        LLaMA3-70b & 0.631 & \textcolor{blue}{0.687} & 0.527 & \textcolor{blue}{0.48} & \textcolor{blue}{0.625} & \textcolor{blue}{0.731} & \textcolor{blue}{0.629} & 0.582 & \textcolor{blue}{0.738} & 0.516 & \textcolor{blue}{0.622} & \textcolor{blue}{0.585} & 0.526 \\
        \noalign{\hrule height 1.5pt}
        \rowcolor{gray!20}\multicolumn{14}{c}{\it{\textbf{Open Source LLMs}}} \\
        \hline
        \rowcolor{yellow}\textbf{Qwen2-7B} & 0.338 & 0.347 & 0.27 & 0.301 & 0.497 & 0.516 & 0.318 & 0.379 & 0.236 & 0.295 & 0.34 & 0.362 & 0.429 \\
        Baichuan2-13B & 0.382 & 0.326 & 0.392 & 0.332 & 0.326 & 0.459 & 0.415 & 0.426 & 0.274 & 0.347 & 0.363 & 0.264 & 0.452 \\
        ChatGLM4-9B & \textcolor{red}{0.143} & \textcolor{red}{0.197} & 0.242 & 0.295 & 0.257 & \textcolor{red}{0.236} & 0.289 & 0.357 & 0.322 & \textcolor{red}{0.268} & \textcolor{red}{0.159} & 0.304 & \textcolor{red}{0.219} \\
        internLM2-20B & 0.429 & 0.357 & 0.32 & 0.259 & 0.238 & 0.326 & 0.325 & 0.373 & 0.391 & 0.305 & 0.283 & 0.362 & 0.345 \\
        \noalign{\hrule height 1.5pt}
        \rowcolor{gray!20}\multicolumn{14}{c}{\it{\textbf{Domain Model}}} \\
        \hline
        ChatMed & \textcolor{blue}{0.739} & 0.659 & 0.631 & & & & & & & & & & \\
        DISC-MedLLM & \textcolor{blue}{0.744} & 0.631 & \textcolor{blue}{0.759} & & & & & & & & & & \\
        FinGPT & & & & \textcolor{blue}{0.625} & 0.593 & 0.639 & & & & & & & \\
        ChatLaw & & & & & & & \textcolor{blue}{0.672} & 0.549 & \textcolor{blue}{0.721} & \textcolor{blue}{0.754} & & & \\
        DISC-LawLLM & & & & & & & 0.625 & \textcolor{blue}{0.587} & 0.648 & \textcolor{blue}{0.805} & & & \\
        MindChat & & & & & & & & & & & \textcolor{blue}{0.542} & 0.583 & \\
        SoulChat & & & & & & & & & & & 0.319 & 0.497 & \\
        MachineMindset(MBTIGPT) & & & & & & & & & & & & & \textcolor{blue}{0.873} \\
        \noalign{\hrule height 1.5pt}
        \rowcolor{gray!20}\multicolumn{14}{c}{\it{\textbf{Task Fine Tuned Model}}} \\
        \hline
        AnyTaskTune-Qwen2-7B-Med & \textcolor{blue}{0.835}{\textcolor{blue}{$\uparrow$}} & \textcolor{blue}{0.751}{\textcolor{blue}{$\uparrow$}} & \textcolor{blue}{0.719}{\textcolor{blue}{$\uparrow$}} & 0.247{\textcolor{red}{$\downarrow$}} & \textcolor{red}{0.234}{\textcolor{red}{$\downarrow$}} & 0.251{\textcolor{red}{$\downarrow$}} & 0.242{\textcolor{red}{$\downarrow$}} & 0.372{\textcolor{red}{$\downarrow$}} & 0.196{\textcolor{red}{$\downarrow$}} & 0.513{\textcolor{blue}{$\uparrow$}} & 0.429{\textcolor{blue}{$\uparrow$}} & 0.358{\textcolor{red}{$\downarrow$}} & 0.326{\textcolor{red}{$\downarrow$}}\\
        AnyTaskTune-Qwen2-7B-Finance & 0.316{\textcolor{red}{$\downarrow$}} & 0.295{\textcolor{red}{$\downarrow$}} & 0.283{\textcolor{blue}{$\uparrow$}} & 0.458{\textcolor{blue}{$\uparrow$}} & \textcolor{blue}{0.739}{\textcolor{blue}{$\uparrow$}} & \textcolor{blue}{0.816}{\textcolor{blue}{$\uparrow$}} & 0.462{\textcolor{blue}{$\uparrow$}} & 0.35{\textcolor{red}{$\downarrow$}} &  0.429{\textcolor{blue}{$\uparrow$}} & 0.391{\textcolor{blue}{$\uparrow$}} & 0.163{\textcolor{red}{$\downarrow$}} & \textcolor{red}{0.193}{\textcolor{red}{$\downarrow$}} & 0.252{\textcolor{red}{$\downarrow$}} \\
        AnyTaskTune-Qwen2-7B-Law & 0.42{\textcolor{blue}{$\uparrow$}} & 0.318{\textcolor{red}{$\downarrow$}} & 0.228{\textcolor{red}{$\downarrow$}} & 0.429{\textcolor{blue}{$\uparrow$}} & 0.523{\textcolor{blue}{$\uparrow$}} & 0.318{\textcolor{red}{$\downarrow$}} & \textcolor{blue}{0.75}{\textcolor{blue}{$\uparrow$}} & \textcolor{blue}{0.743}{\textcolor{blue}{$\uparrow$}} & \textcolor{blue}{0.793}{\textcolor{blue}{$\uparrow$}} & \textcolor{blue}{0.798}{\textcolor{blue}{$\uparrow$}} & 0.451{\textcolor{blue}{$\uparrow$}} & 0.392{\textcolor{red}{$\downarrow$}} & 0.392{\textcolor{red}{$\downarrow$}}\\
        AnyTaskTune-Qwen2-7B-Psychology & 0.195{\textcolor{red}{$\downarrow$}} & 0.228{\textcolor{red}{$\downarrow$}} & \textcolor{red}{0.217}{\textcolor{red}{$\downarrow$}} & 0.232{\textcolor{red}{$\downarrow$}} & 0.418{\textcolor{red}{$\downarrow$}} & 0.385{\textcolor{red}{$\downarrow$}} & 0.314{\textcolor{red}{$\downarrow$}} & \textcolor{red}{0.252}{\textcolor{red}{$\downarrow$}} & \textcolor{red}{0.158}{\textcolor{red}{$\downarrow$}} & 0.329{\textcolor{red}{$\downarrow$}} & 0.469{\textcolor{blue}{$\uparrow$}} & \textcolor{blue}{0.659}{\textcolor{blue}{$\uparrow$}} & 0.572{\textcolor{blue}{$\uparrow$}}  \\
        AnyTaskTune-Qwen2-7B-RolePlay & 0.275{\textcolor{red}{$\downarrow$}} & 0.284{\textcolor{red}{$\downarrow$}} & 0.315{\textcolor{blue}{$\uparrow$}} & \textcolor{red}{0.219}{\textcolor{red}{$\downarrow$}} & 0.238{\textcolor{red}{$\downarrow$}} & 0.316{\textcolor{red}{$\downarrow$}} & 0.413{\textcolor{blue}{$\uparrow$}} & 0.294{\textcolor{red}{$\downarrow$}} & 0.429{\textcolor{blue}{$\uparrow$}} & 0.362{\textcolor{blue}{$\uparrow$}} & 0.459{\textcolor{blue}{$\uparrow$}} & 0.592{\textcolor{blue}{$\uparrow$}} & \textcolor{blue}{0.711}{\textcolor{blue}{$\uparrow$}} \\
        \bottomrule[1.5pt]
    \end{tabular}
\end{table}

\subsection{Detailed Comparative Analysis}

\subsubsection{Impact Across Domains}
\textbf{Medical Domain:} The AnyTaskTune-Qwen2-7B-Med model showcased exceptional performance in medical sub-tasks, notably achieving a F1 score of 0.835 in M1, which is a substantial improvement over the base Qwen2-7B model's score of 0.338. This precision underscores its reliability in handling medical data. However, when this model was tested in finance and law domains, its performance dropped to 0.247 and 0.242 respectively, illustrating a significant decline due to the model's parameters being highly specialized to medical data.

\textbf{Financial Domain:} AnyTaskTune-Qwen2-7B-Finance performed excellently in financial tasks, especially F2 and F3, with scores of 0.739 and 0.816, outperforming the base model's scores of 0.497 and 0.516 in these tasks. Applying this model to legal tasks resulted in moderate performance drops (from 0.462 in legal tasks down from 0.458 in finance tasks), likely due to some overlap in structured data and terminological precision between the finance and legal fields. However, performance in the medical domain was notably poorer, confirming the domain-specific nature of the fine-tuning.

\textbf{Legal Domain:} The AnyTaskTune-Qwen2-7B-Law excelled in legal tasks with scores of 0.75, 0.743, 0.793, and 0.798 across L1, L2, L3, and L4. Its performance in the finance domain was competitive, with scores slightly lower but still respectable. However, when applied to the medical and psychological domains, the performance was limited, further highlighting the challenges of cross-domain applicability due to different knowledge bases and data characteristics.

\textbf{Psychology and Role-Play Domains}
The AnyTaskTune-Qwen2-7B-Psychology model showed good results in psychology, particularly in the P2 task with a score of 0.659. The nature of psychological tasks, which often involve complex language and emotional analysis, allows this model to perform reasonably well in role-play scenarios as well, evidenced by a score of 0.572 in the MBTI task.

\textbf{Role-Play Domain:} Similar to the psychology model, AnyTaskTune-Qwen2-7B-RolePlay demonstrated good performance due to overlapping requirements in emotional and language processing between the two fields.

\subsubsection{Task Adaptability and Generalizability}
While domain-specific fine-tuning led to exceptional performance within respective fields, it generally resulted in reduced effectiveness when models were applied across unrelated domains. For instance, while AnyTaskTune-Qwen2-7B-Med excelled in medical tasks, its application in non-medical fields showed significant performance reductions, a clear indication of the trade-offs involved in specialized fine-tuning.

\textbf{Comparison with Base Model:} The original Qwen2-7B model, without fine-tuning, generally showed lower performance across all specific tasks compared to the fine-tuned AnyTaskTune models. This highlights the effectiveness of task-specific fine-tuning in enhancing model capabilities, although it restricts the model’s flexibility across various domains.

\subsection{Sub-task Details}

The specific types of sub-tasks evaluated in this experiment will be detailed in the accompanying GitHub repository. This repository will provide comprehensive descriptions and examples of each sub-task, offering valuable insights for further research and application development.

\begin{figure}[htp]
 \centering
 \includegraphics[width=1.0\columnwidth]{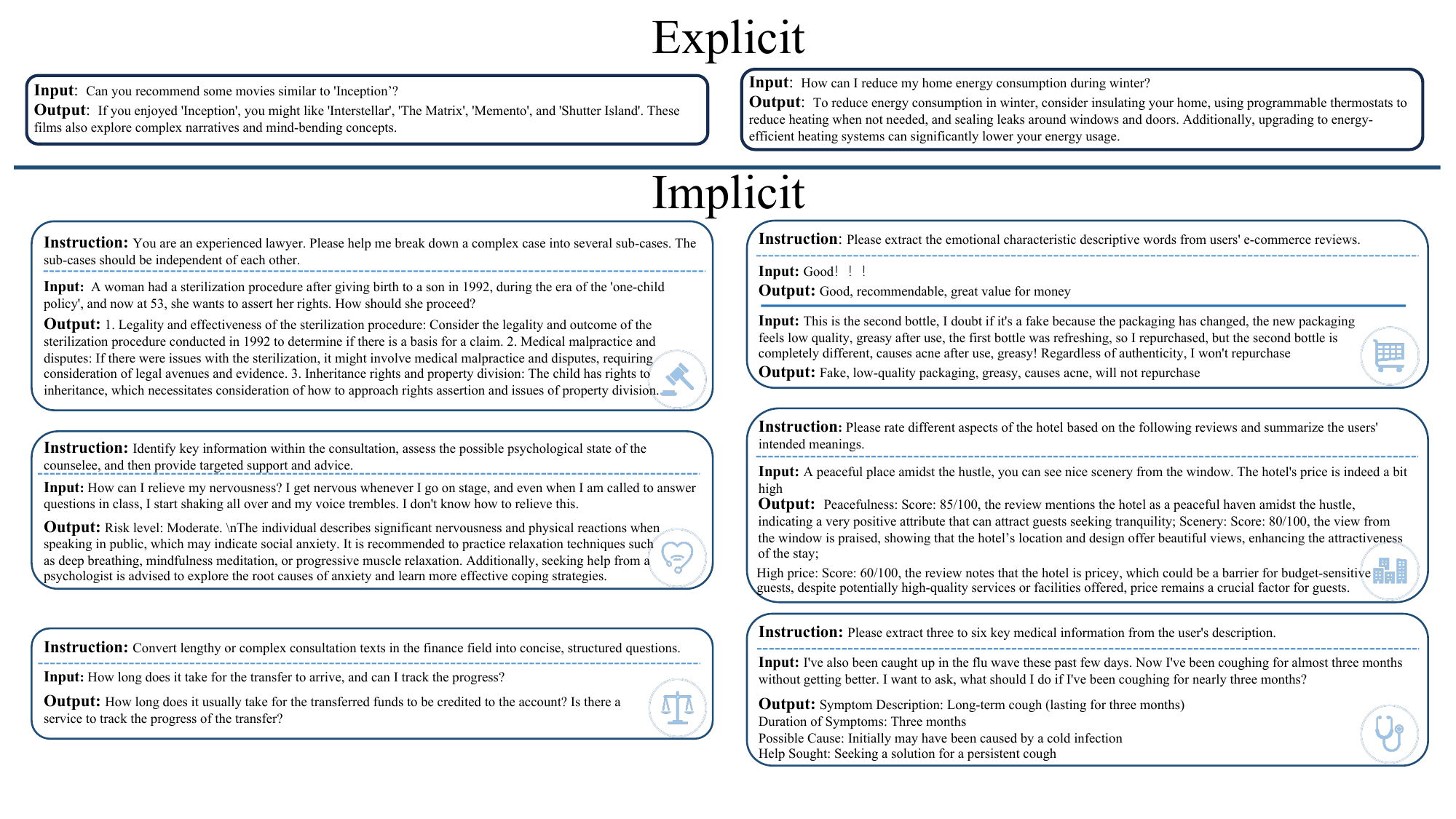}
 \caption{Task examples}
 \label{fig:motivation1}
\end{figure}

\subsection{Analysis}

This analysis confirms that while AnyTaskTune can dramatically improve performance on specific tasks, its cross-domain applicability is limited. This highlights the importance of task-specific datasets in fine-tuning processes, ensuring that models are not only highly effective but also tailored to the particular characteristics and requirements of their intended operational domains.

\vspace{-0.5em}
\section{Conclusion}
\vspace{-0.5em}

In this work, we introduced AnyTaskTune, a novel fine-tuning paradigm specifically designed for Explicit Data Sets. Our approach addresses the limitations of traditional general models and domain-specific models by emphasizing precision and specificity, which are critical for real-world applications. By leveraging our extensive experience in various vertical domains, we developed and utilized over twenty bilingual Explicit Data Sets, enabling us to fine-tune models for specific tasks with high accuracy and efficiency.

Our experiments demonstrated that AnyTaskTune significantly outperforms both closed-source and open-source LLMs, as well as existing domain-specific models, across a range of tasks in finance, healthcare, law, psychology, and role-play. The results showcased the superior performance of AnyTaskTune in handling complex and nuanced tasks, thereby validating the effectiveness of task-specific fine-tuning.

Moreover, we highlighted the practical implications of AnyTaskTune for businesses and organizations. By defining clear and directive input-output pairs through Explicit Data Sets, we enable precise model tailoring, which not only enhances task performance but also improves the model's applicability and efficiency in specific operational contexts. Our methodology ensures that models are not just broadly capable but are finely tuned to meet the exact needs of various business scenarios.

We also introduced ~\href{https://datatager.com}{DataTager}, our foundational data synthesis model, which forms the basis for generating high-quality datasets used in AnyTaskTune. While we have not open-sourced DataTager-LLM, we have provided comprehensive documentation and resources in our accompanying GitHub repository, detailing the sub-tasks and datasets used in our experiments. This transparency aims to foster further research and development in the community.

In conclusion, AnyTaskTune represents a significant advancement in the development and deployment of Large Language Models. By shifting the focus from general capabilities to specialized, task-oriented model training, AnyTaskTune offers a robust and efficient solution for industry-specific applications. This strategic evolution in model fine-tuning not only enhances performance but also ensures that LLMs are more relevant and valuable in real-world business contexts.

We believe that AnyTaskTune will pave the way for more precise and effective applications of AI in various domains, ultimately contributing to the broader adoption and integration of AI technologies in everyday business operations.

\newpage
\section*{Authorship, Credit Attribution, and Acknowledgements}

Please cite this work as ``DataTager(2024)''.

\vspace{10pt}

\textbf{Authorship and Contributions}

\vspace{10pt}

\textbf{Paper Writing}

\begin{itemize}
    \item \textbf{Jiaxi Cui} - Founder, Provided the main ideas, data construction, model training, paper writing.
    \item \textbf{Wentao Zhang} - Organized ideas and outline, and contributed to paper writing.
\end{itemize}

\vspace{10pt}

\textbf{Engineering}

\begin{itemize}
    \item \textbf{Xudong Tong} - Software engineering at \href{https://datatager.com}{DataTager}.
    \item \textbf{Zhenwei Zhang} - Tencent - Software engineering at \href{https://datatager.com}{DataTager}.
\end{itemize}

\vspace{10pt}

\textbf{Other Contributions}

\begin{itemize}
    \item \textbf{Jing Tang} - Huazhong University of Science and Technology - Responsible for external presentations and resource acquisition.
    \item \textbf{Amie} - PublicAI \& Beihang University - Web3 and external resource acquisition
    \item \textbf{Jing Wen} - Provided industry insights.
    \item \textbf{Rongsheng Wang} - Qiyuan.Tech - Provided assistance in the medical field.
    \item \textbf{Pengfei Wu} - Peking University \& Tencent - Offered partial assistance.
\end{itemize}

\vspace{10pt}

\textbf{Acknowledgements}

We would like to express our sincere gratitude to the following individuals and organizations for their invaluable support and contributions to this project:

\begin{itemize}
    \item Modelscope's Chen Cheng and HuggingFace's Tiezhen Wang for their tremendous support.
    \item \href{https://publicai.io/}{PublicAI}, our Web3 partner, for providing data support.
    \item Sci-Learning, our channel partner, for reaching student groups.
    \item Yuan-Group for supporting the free exploration of early ideas.
    \item Fudan NLP team for their related research contributions.
    \item The open-source community and contributors to the various software libraries used in this project. Your dedication and hard work are deeply appreciated.
\end{itemize}

\vspace{10pt}

\textbf{Special Thanks}

A special thanks to the open-source community and contributors to the various software libraries used in this project. Your dedication and hard work are deeply appreciated.

\vspace{10pt}

\textbf{Contact Information}

For further information or inquiries, please contact us at \href{mailto:contact@datatager.com}{report@datatager.com}.

\vspace{20pt}

\noindent\textit{Thank you for your interest and support in our work.}

\begin{flushright}
    \textbf{The DataTager Team} \\
    \today
\end{flushright}

\bibliography{example_paper}
\bibliographystyle{abbrv}

\newpage
\appendix
\onecolumn

\end{document}